\documentclass{article}
\usepackage{ijcai20}

\usepackage{times}
\usepackage{soul}
\usepackage[utf8]{inputenc}
\usepackage[small]{caption}
\usepackage{graphicx}
\usepackage{amsmath}
\usepackage{amsthm}
\usepackage{booktabs}
\usepackage{algorithm}
\usepackage{algorithmic}

\usepackage{amsmath,amsfonts,amssymb }
\usepackage{multirow}
\usepackage{algorithm}
\usepackage{algorithmic}
\usepackage{subfigure}
\usepackage{amsthm}
\usepackage{enumitem}
\usepackage{flushend}

\graphicspath{{./img/}}

\title{Multi-Level Generative Models for Partial Label Learning \\ with Non-random Label Noise}

\author{%
	Yan Yan$^{1,2}$ 
	\and
  	Yuhong Guo$^1$\\
\affiliations
  $^1$School of Computer Science,
  Carleton University, Canada\\
 $^2$School of Computer Science and Engineering,
Northwestern Polytechnical University, China 
}

\begin{document}

\maketitle

\begin{abstract}
Partial label (PL) learning tackles the problem where each training instance is associated with 
a set of candidate labels that include both the true label 
and irrelevant noise labels.
	In this paper, we propose a novel multi-level generative model for partial label learning (MGPLL),
which tackles
the problem by learning both a label level adversarial generator and a feature level adversarial generator
under a bi-directional mapping framework between the label vectors and the data samples.
Specifically, MGPLL uses a conditional noise label generation network to model the non-random noise labels 
and perform label denoising,
	and uses a multi-class predictor
to map the training instances to the denoised label vectors,
while a conditional data feature generator is used to form an inverse mapping
from the denoised label vectors to data samples. 
Both the noise label generator and the data feature generator are learned in an adversarial manner
to match the observed candidate labels and data features respectively. 
Extensive experiments are conducted on synthesized and real-world partial label datasets. 
The proposed approach
demonstrates the state-of-the-art performance for partial label learning.

\end{abstract}

\section{Introduction}

Partial label (PL) learning is a weakly supervised learning problem with ambiguous labels \cite{hullermeier2006learning,zeng2013learning}, where each training instance is assigned a set of candidate labels,
among which only one is the true label.
Since it is typically difficult and costly to annotate instances precisely, the task of partial label learning naturally 
arises in
many real-world learning scenarios, including automatic face naming \cite{hullermeier2006learning,zeng2013learning}, 
and web mining \cite{luo2010learning}. 

As the true label information is hidden in the candidate label set, 
the main challenge of PL lies in identifying the ground truth labels from the candidate noise labels.
Intuitively, one basic strategy to tackle partial label learning is performing label disambiguation.
There are two main groups of disambiguation-based PL approaches:
the average-based disambiguation approaches 
and the identification-based approaches.
For averaging-based disambiguation, each candidate label is treated equally in model induction and the final prediction is made by
averaging the modeling outputs of all the candidate labels \cite{hullermeier2006learning,cour2011learning,zhang2016partial}.
Without differentiating the informative true labels from the noise irrelevant labels, 
such simple averaging methods in general cannot produce satisfactory performance. 
Hence recent studies are mostly focused on identification-based disambiguation methods. 
Many identification-based disambiguation methods treat the ground-truth labels as latent variables 
and identify the true labels by employing 
iterative label refining procedures \cite{jin2003learning,zhang2015solving,tang2017confidence}.
For example, 
the work in \cite{feng2018leveraging} tries to estimate the latent label distribution with iterative label propagations
and then induce a prediction model by fitting the learned latent label distribution.
Another work in \cite{Feng2019Retraining} exploits a self-training strategy to induce
label confidence values and learn classifiers in an alternative manner
by minimizing the squared loss between the model predictions and the learned label confidence matrix.
However, these methods suffer from the cumulative errors induced in either the separate label distribution estimation steps
or the error-prone label confidence estimation process. 
Moreover, all these methods have a common drawback:
they automatically assumed random noise in the label space -- that is, they assume the 
noise labels are randomly distributed. 
However, in real world problems the appearance of noise labels are usually dependent on the target true labels.
For example, when the object contained in an image is a ``computer", a noise label ``TV" could be added due
to a recognition mistake or image ambiguity, but it is less likely 
to annotate the object as ``lamp" or ``curtain",
while the probability of getting noise labels such as ``tree" or ``bike" is even smaller.

In this paper, we propose a novel multi-level adversarial generative model, MGPLL, for partial label learning. 
The MGPLL model comprises of conditional data generators at both the label level and feature level.
The noise label generator directly models non-random appearances of noise labels
conditioning on the true label by adversarially matching the candidate label observations,
while the data feature generator models the data samples conditioning on the corresponding true labels
by adversarially matching the observed data sample distribution. 
Moreover, a prediction network is incorporated to predict the denoised true label of each instance
from its input features, 
which forms inverse mappings between labels and features, together with the data feature generator.
The learning of the overall model corresponds to a minimax adversarial game,
which simultaneously identifies true labels of the training instances from 
both the observed data features and the observed candidate labels, 
while inducing accurate prediction networks that map input feature vectors to (denoised) true label vectors. 
To the best of our knowledge, this is the first work that exploits 
multi-level generative models to model non-random noise labels for partial label learning.
We conduct extensive experiments on real-world and synthesized PL datasets. 
The empirical results show the proposed MGPLL achieves the state-of-the-art PL performance.

\section{Related Work}

Partial label learning is a weakly supervised classification problem in many real-world domains, 
where 
the true label of each training instance is hidden within a given candidate label set.
The PL setting is different from the noise label learning, 
where the ground-truth labels on some instances are replaced by noise labels.
The key for PL learning lies in how to disambiguate the candidate labels.
Existing disambiguation-based approaches mainly follow two strategies:
the average-based strategy and the identification-based strategy.

The average-based strategy assumes that each candidate label contributes equally to model training,
and then averages the outputs of all the candidate labels for final prediction.
Following such a strategy, the discriminative learning methods \cite{cour2011learning,zhang2016partial}
distinguish the averaged model outputs based on all the candidate labels 
from the averaged outputs based on all the non-candidate labels.
The instance-based learning methods \cite{hullermeier2006learning,gong2018regularization}
on the other hand predict the label for a test instance 
by averaging the candidate labels from its neighbors.
The simple average-based strategy however cannot produce satisfactory performance since
it fails to take the difference among the candidate labels into account.

By considering the differences between candidate labels,
the identification-based strategy has gained increasing attention due to its effectiveness of
handling the candidate labels with discrimination.
Many existing approaches following this strategy take the ground-truth labels as latent variables. 
Then the latent variables and model parameters are refined via EM procedures which optimize the objective function
based on a maximum likelihood criterion \cite{jin2003learning} or a maximum margin criterion \cite{nguyen2008classification}.
Recently, some researchers proposed to learn the label confidence value of each candidate label by
leveraging the topological information in feature space and achieved some promising results
\cite{zhang2015solving,feng2018leveraging,Xu2019Enhancement,feng2019partial,wang2019adaptive}.
One previous work in \cite{feng2018leveraging} attempts to estimate the latent label distribution
using iterative label propagation along the topological information extracted based on the local consistency assumption;
i.e., nearby instances are supposed to have similar label distributions.
Then, a prediction model is induced using the learned latent label distribution.
However, the latent label distribution estimation can be impaired by the cumulative error induced in the propagation process,
which can consequently degrade the partial label learning performance.
Another work in \cite{Feng2019Retraining} tries to refine the label confidence values with a self-training
strategy and induce the prediction model over the refined label confidence scores by alternative optimization.
However, due to the nature of alternative optimization,
the estimation error on confidence values can 
negatively impact
the coupled partial label predictor,
Moreover, all these existing methods have assumed random label noise by default, 
which however does not hold in many real world learning scenarios. 
This paper presents the first work that explicitly model non-random noise labels
for partial label learning.

\section{The Proposed Approach}

\begin{figure}[t]
\centering
\includegraphics[width=9cm]{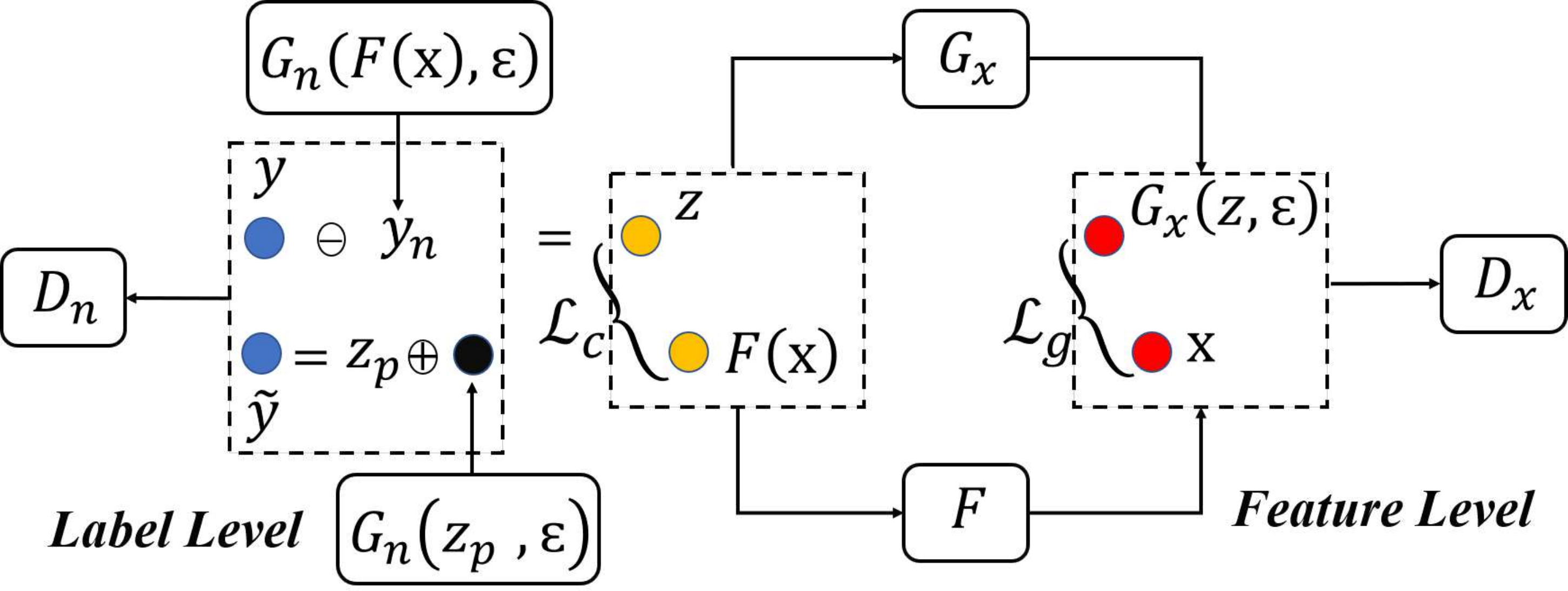}
\caption{The proposed MGPLL model
comprises five component networks:
the conditional noise label generator, $G_n$,
which models noise labels conditioning on the truth label at the label level;
the feature level conditional data generator, $G_x$,
generates data samples conditioning on
the denoised label vectors; 
the discriminator, $D_n$,
	tries to maximally separate the generated candidate label vectors from the observed real ones;
the discriminator, $D_x$,
tries to maximally separate the generated samples from the real data; 
the prediction network, $F$,
predicts the denoised label vector for each training sample,
and together with $G_x$ forms bi-directional maps between the label and feature spaces.
	}
\label{MGPLL}
\end{figure}

Given a partial label training set $S = \{(\mathbf{x}_i, \mathbf{y}_i)\}\operatorname{}\limits_{i=1}^n $,
where $\mathbf{x}_i \in \mathbb{R}^d$ is a \emph{d}-dimensional feature vector for the $i$-th instance,
and $\mathbf{y}_i \in \{0, 1\}^L$ denotes the candidate label indicator vector associated with $\mathbf{x}_i$,
which has multiple 1 values corresponding to the ground-truth label and the noise labels,
the task is to learn a good multi-class prediction model. 
In real world scenarios, the irrelevant noise labels are typically not presented in a random manner, 
but rather correlated with the ground-truth label. 
In this section, we present a novel multi-level generative model for partial label learning, MGPLL.
The model is illustrated in Figure \ref{MGPLL}. 
It models non-random noise labels using an adversarial conditional noise label generator
$G_n$ with a corresponding discriminator $D_n$,
and builds connections between the denoised label vectors and instance features 
using a label-conditioned feature level generator $G_x$ and a label prediction network $F$. 
The overall model learning problem corresponds to a minimax adversarial game, 
which conducts multi-level generator learning by matching the observed data
in both the feature and label spaces, while boosting 
the correspondence relationships between features and labels
to induce an accurate multi-class prediction model.
Below we present the details of the two level generations, the prediction network,
and the overall learning problem.

\subsection{Conditional Noise Label Generation}

The key challenge of partial label learning 
lies in the fact that ground-truth label is hidden among noise labels 
in the given candidate label set. 
As aforementioned, in real world partial label learning problems, 
the appearances of noise labels are typically not random,
but rather correlated with the ground-truth labels. 
Hence we propose a conditional noise label generation model
to model the appearances of the target-label dependent noise labels and 
match the observed candidate label distribution in the training data through adversarial learning. 

Specifically, 
given a noise value sampled from a uniform distribution $\epsilon \sim P_{\epsilon}$
and a one-hot label indicator vector $\mathbf{z}$ 
sampled from a multinomial distribution $P_{\mathbf{z}}$,
we use a noise label generator 
$G_{n}(\mathbf{z}, \epsilon)$ to generate 
a noise label vector conditioning on the true label $\mathbf{z}$, 
which can be combined with ${\bf z}$ 
to form a generated candidate label vector $\mathbf{\tilde{y}}$,
such that
$\mathbf{\tilde{y}}
=G_{n}(\mathbf{z}, \epsilon)\oplus {\bf z} 
=\min(G_{n}(\mathbf{z}, \epsilon)+ {\bf z},1)$. 
We then adopt the adversarial learning principle to learn such a noise label 
generation model
by introducing a discriminator 
$D_{n}({\bf y})$,
which is a two-class classifier and 
predicts 
how likely
a given label vector ${\bf y}$
comes from the real data 
instead of generated data. 
By adopting the adversarial loss of the 
Wasserstein Generative Adversarial Network (WGAN)
\cite{pmlr-v70-arjovsky17a},
our adversarial learning problem can be formulated as
the following minimax optimization problem:
\begin{align}
 \label{advn_loss}
\min_{G_n}\max_{D_n} & \quad \mathcal{L}^n_{adv}(G_n, D_n) = 
\\	
&\mathbb{E}_{(\mathbf{x}_i, \mathbf{y}_i) \sim S}D_n(\mathbf{y}_i)
            - \mathbb{E}_{\mathbf{z} \sim P_{\mathbf{z}} \atop \epsilon \sim P_{\epsilon}}
	    D_n(G_n(\mathbf{z}, \epsilon)\oplus{\bf z})
\nonumber	    
\end{align}
Here the discriminator $D_n$ 
attempts to maximally distinguish the generated candidate label vectors
from the observed candidate label indicator vectors in the real training data,
while 
the generator $G_n$ tries to 
generate noise label vectors and hence candidate label vectors that are similar to the real data
in order to maximally confuse the discriminator $D_n$.
By playing a minimax game between the generator $G_n$ and the discriminator $D_n$, 
the adversarial learning is expected to induce a generator $G_n^*$ such that
the generated candidate label distribution can match the observed candidate label distribution
in the training data~\cite{GAN}. 
We adopted the training loss of WGAN here, 
as WGAN can
overcome the mode collapse problem 
and 
have improved learning stability 
comparing to the standard GAN
\cite{pmlr-v70-arjovsky17a}.

Note although the proposed generator $G_n$ is designed to model true-label dependent noise labels,
it can be easily modified to model random noise label distributions by 
simply dropping the label vector input to have $G_n(\epsilon)$.

\subsection{Prediction Network}

The ultimate goal of partial label learning is to learn an accurate prediction network $F$. 
To train a good predictor, we need to obtain denoised labels on the training data. 
For a candidate label indicator vector $\bf{y}$,  if the noise label indicator vector ${\bf y}_n$ 
is given, one can simply perform label denoising as follows 
to obtain the corresponding true label vector ${\bf z}$:
\begin{align}
	{\bf z} = {\bf y}\ominus {\bf y}_n = \max({\bf y-y_n}, 0)
\end{align}
Here the operator ``$\ominus$" is introduced to generalize the standard minus ``$-$" operator
into the non-ideal case, 
where the noise label indicator vector ${\bf y}_n$ is not properly contained
in the candidate label indicator vector. 

The generator $G_n$ presented in the previous section
provides a mechanism to generate noise labels and denoise candidate label sets, 
but requires true target label vector as input.
We can use the outputs of the prediction network $F$ to approximate the 
target true label vectors of the training data for candidate label denoising purpose
with $G_n$, while using the denoised labels as the prediction target for $F$.
Specifically, 
with the noise label generator $G_n$ and predictor $F$, we can perform partial label learning
by minimizing the following classification loss on the training data $S$:
\begin{align}
    \label{classify_loss}
	\operatorname*{min}\limits_{F, G_{{n}}}\quad &
        \mathcal{L}_{c}(F, G_{{n}}) = \\
	& \mathbb{E}_{
	\epsilon \sim P_\epsilon\quad \atop 
(\mathbf{x}_i, \mathbf{y}_i) \sim S  
	}\;
	\ell_c\big(F(\mathbf{x}_i),\; \mathbf{y}_i\ominus G_n(F(\mathbf{x}_i),\epsilon)\big)
\nonumber	
\end{align}
Although in the ideal case, the output vectors of $G_n$ and $F$ would be indicator label vectors, 
it is error-prone and difficult for neural networks to output discrete values. 
To pursue more reliable predictions and avoid overconfident outputs,
$G_{{n}}$ and $F$ predict the probability of
each class label being a noise label and ground-truth label respectively.
Hence the loss function $\ell_c(\cdot,\cdot)$ in Eq.(\ref{classify_loss}) above 
denotes a mean square error loss
between the predicted probability of each label being the true label (through $F$)
and its confidence of being a ground-truth label (through $G_n$).

\subsection{Conditional Feature Level Data Generation}

With the noise label generation model and the prediction network above, 
the observed training data in both the label and feature spaces are exploited to recognize 
the true labels and induce good prediction models. 
Next, we incorporate a conditional data generator $G_x({\bf z},\epsilon)$ at the feature level
to map (denoised) label vectors in the label space into instances in the feature space, 
aiming to further strengthen 
the mapping relations between data samples and the corresponding labels,
enhance label denoising and hence improve partial label learning performance. 
Specifically,
given a noise value $\epsilon$ sampled from a uniform distribution $P_{\epsilon}$
and a one-hot label vector $\mathbf{z}$ sampled from a multinomial distribution $P_{\mathbf{z}}$,
$G_{{x}}(\mathbf{z}, \epsilon)$ generates an instance in the feature space 
that is corresponding to label ${\bf z}$. 
Hence given the training label vectors in $S$ denoised with $G_n$, 
the data generator $G_x$ is expected to regenerate the corresponding training instances in the feature space.
This assumption can be captured using the following generation loss:
\begin{align}
    \label{rec_loss}
&\mathcal{L}_{g}(F, G_{{n}}, G_{{x}}) = 
        \mathbb{E}_{(\mathbf{x}_i, \mathbf{y}_i) \sim S \atop \epsilon_1, \epsilon_2 \sim P_{\epsilon}}
	\; \ell_{g} \big(G_{{x}}({\bf z}_i,\epsilon_2), \mathbf{x}_i\big)
\end{align}
where ${\bf z}_i = {\bf y}_i\ominus G_n(F({\bf x}_i),\epsilon_1)$ denotes the denoised label vector
for the $i$-th training instance, and 
$\ell_{g}(\cdot,\cdot)$ is a mean square error loss function. 

Moreover, by introducing a discriminator $D_x({\bf x})$,
which predicts how likely a given instance ${\bf x}$ is real, 
we can deploy an adversarial learning scheme to learn the generator $G_x$
through the following minimax optimization problem with the WGAN loss:
\begin{align}
    \label{advx_loss}
     \operatorname*{min} \limits_{G_{{x}}} \operatorname*{max}\limits_{D_{{x}}} \;\;
	&\mathcal{L}^x_{adv}( G_{{x}}, D_{{x}}) = \\
        &\mathbb{E}_{(\mathbf{x}_i, \mathbf{y}_i) \sim S}D_{{x}}(\mathbf{x}_i)
            - \mathbb{E}_{\mathbf{z} \sim P_{\mathbf{z}} \atop \epsilon \sim P_{\epsilon}}
            D_{{x}}(G_{{x}}(\mathbf{z}, \epsilon))
\nonumber   
\end{align}
By playing a minimax game between $G_x$ and $D_x$, 
this adversarial learning is expected to induce a generator $G^*_x$ that can generate samples
with the same distribution as the observed training instances. 
Hence the mapping relation from label vectors to samples induced by $G^*_x$
can also hold on the real training data,
and should be consistent with the inverse mapping from samples to label vectors
through the prediction network. 
Therefore, we can further consider an auxiliary classification loss on the generated data:
\begin{align}
    \label{aux_loss}
      \mathcal{L}_{c^{\prime}}(F, G_{{x}}) =
        \mathbb{E}_{\mathbf{z} \sim P_{\mathbf{z}} \atop \epsilon \sim P_{\epsilon}}\;\,
	    \ell_{c^{\prime}} \big(F(G_{{x}}(\mathbf{z}, \epsilon)), \mathbf{z}\big)
\end{align}
where $\ell_{c^{\prime}}(\cdot,\cdot)$ can be a cross-entropy loss between the label prediction probability vector
and the sampled true label indicator vector.

\subsection{Learning the MGPLL Model}
By integrating the classification loss in Eq.(\ref{classify_loss}), 
the adversarial losses in Eq.(\ref{advn_loss}) and 
Eq.(\ref{advx_loss}), the generation loss in Eq.(\ref{rec_loss}) and
the auxiliary classification loss in Eq.(\ref{aux_loss}) together,
MGPLL learning can be formulated as the following min-max optimization problem:
\begin{align}
        &\operatorname*{min} \limits_{G_{{n}}, G_{{x}}, F} \ \operatorname*{max}\limits_{D_{{n}}, D_{{x}}}
         \mathcal{L}_{c}(F, G_{{n}}) + \mathcal{L}^n_{adv}(G_{{n}}, D_{{n}}) + \nonumber\\  
			     &\quad\;\alpha \mathcal{L}^x_{adv}(G_{{x}}, D_{{x}})  
                             +\beta \mathcal{L}_{g}(F, G_{{n}}, G_{{x}})
                             +\gamma \mathcal{L}_{c^{\prime}}(F, G_{{x}})
    \label{total_loss}
\end{align}
where $\alpha$, $\beta$ and $\gamma$ are trade-off hyperparameters. 
The learning of the overall model corresponds to a minimax adversarial game.
We develop a batch-based stochastic gradient descent algorithm to solve it 
by conducting minimization over \{$G_n, G_x, F$\} and maximization over \{$D_n, D_x$\} alternatively. 
The overall training algorithm is outlined in Algorithm \ref{alg_1}.

\begin{algorithm}[t!]
\caption{Minibatch stochastic gradient descent.}
\label{alg_1}
\textbf{Input}: $S:$ the PL training set;
                $\alpha, \beta, \gamma$: the trade-off hyperparameters;
                $c$: the clipping parameter;
                $m$: minibatch size. \\
\vspace{-0.2cm}
\begin{algorithmic}[0] 
\STATE \textbf{for} number of training iterations \textbf{do} \\
   \STATE \hspace{0.3cm} Sample a minibatch $B$ of $m$ samples from $S$.
	\STATE \hspace{0.3cm} Sample a $m$ noise values $\{\epsilon_i\}$ from a prior $ P(\epsilon)$. \\
   \STATE \hspace{0.3cm} Sample $m$ label vectors $\{\mathbf{z}_i\}$ from a prior $P_{\mathbf{z}}$. \\
   \STATE \hspace{0.3cm} Update $D_{{n}}, D_{{x}}$ by ascending their stochastic gradients:\\
   \STATE \hspace{0.5cm} 
	{\small
   	\begin{align*}
	\nabla_{\Theta_{\tiny D_{{n}}, D_{{x}}}}\frac{1}{m} \operatorname*{\sum}\limits_{i=1}^{m}
        \left \{\!\!\!\!
		\begin{array}{l}	
		\Big(D_{{n}}(\mathbf{y}_i)
	    - D_{{n}}(G_{{n}}(\mathbf{z}_i, \epsilon_i)\oplus{\bf z}_i)\Big)\\
            + \alpha \Big(D_{{x}}(\mathbf{x}_i)
            - D_{{x}}(G_{{x}}(\mathbf{z}_i, \epsilon_i))\Big)
		\end{array}
            \!\!\!\!\!\!\right \}
   	\end{align*}
	}
	\vspace{.2cm}	    
   \STATE \hspace{0.3cm} $\Theta_{D_{\mathbf{n}}, D_{\mathbf{x}}} \gets \mathrm{clip}(\Theta_{D_{\mathbf{n}}, D_{\mathbf{x}}}, -c, c)$
	(WGAN adjustment)\\
	\vspace{.2cm}	    
	\STATE \hspace{0.3cm} Sample $m$ noise values $\{\overline \epsilon_i\}$ from a prior $ P(\epsilon)$. \\
   \STATE \hspace{0.3cm} Update $G_{\mathbf{n}}, G_{\mathbf{x}}, F$ by stochastic gradient descent:
   \STATE \hspace{.5cm} 
	{\small
   	\begin{align*}
	\nabla_{\Theta_{\tiny G_{{n}}, G_{{x}},F}} \frac{1}{m}\! \operatorname*{\sum}\limits_{i=1}^{m} 
	\!\!
	\left\{\!\!\!\!
		\begin{array}{l}	
		\ell_c (F(\mathbf{x}_i), \mathbf{y}_i\ominus  G_{{n}}(F(\mathbf{x}_i), \epsilon_i)) \\
			- D_{{n}}(G_{{n}}(\mathbf{z}_i, \epsilon_i)\oplus {\bf z}_i) \\
   - \alpha D_{{x}}(G_{{x}}(\mathbf{z}_i, \overline\epsilon_i))+ \\
\beta\ell_{g} \big(G_{{x}}\big(\mathbf{y}_i \ominus G_{{n}}(F(\mathbf{x}_i), \epsilon_i),\overline \epsilon_i\big), \mathbf{x}_i\big)\\
   + \gamma \ell_{c^{\prime}} F(G_{{x}}(\mathbf{z}_i, \overline\epsilon_i), \mathbf{z}_i) 
		\end{array}
	\!\!\!\!	
	\right\}
   	\end{align*}
	}
\STATE \textbf{end for} \\
\end{algorithmic}
\end{algorithm}

\begin{figure*}[t!]
\centering
\begin{minipage}[b]{0.8\textwidth}
\includegraphics[width=1\textwidth]{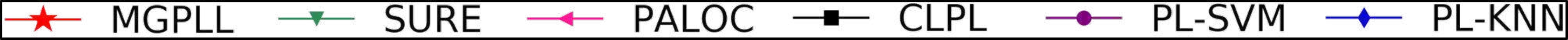}
\end{minipage}
\subfigure[ecoli]{
\label{fig:example1}
\includegraphics[width=.385\columnwidth]{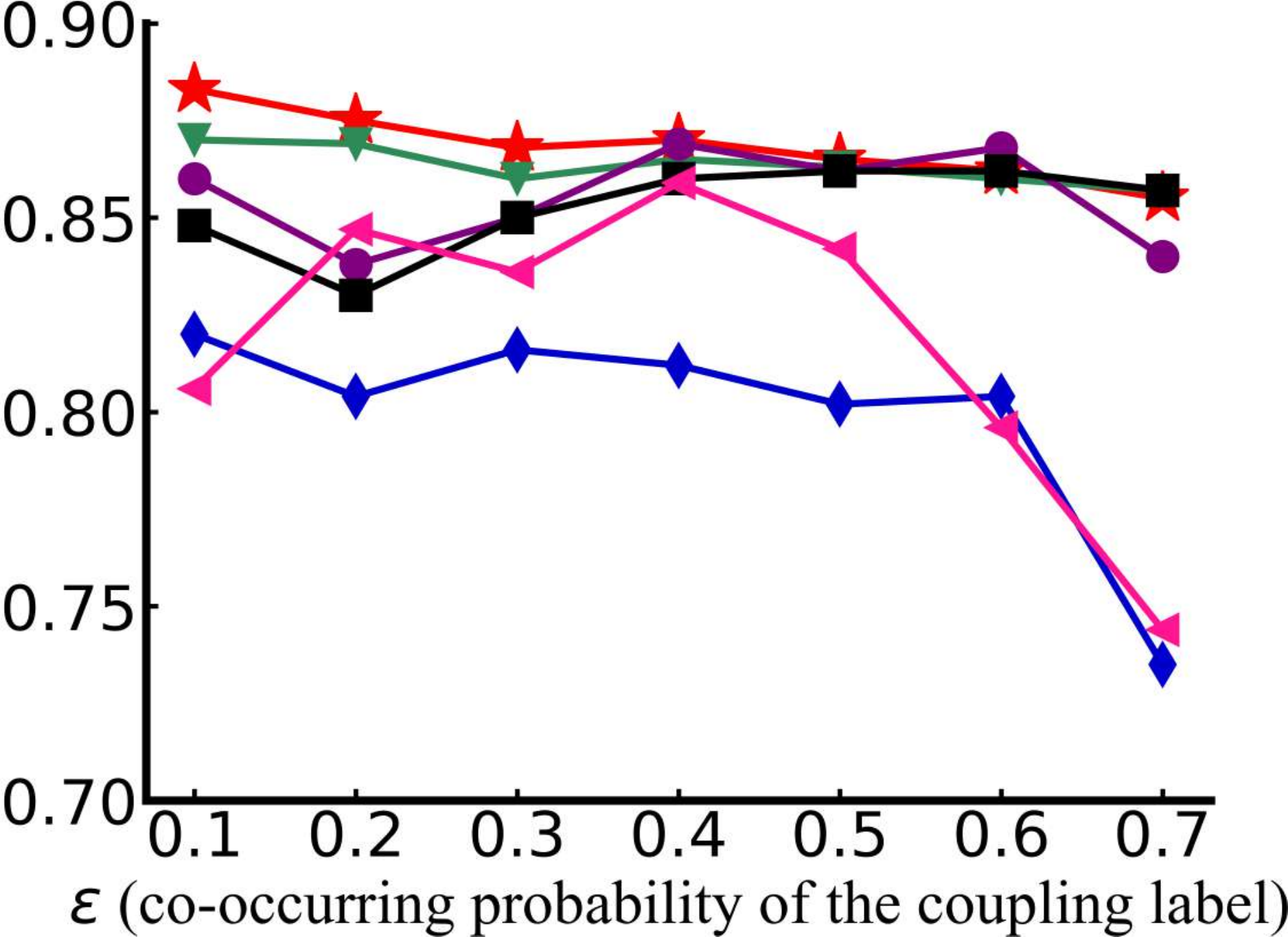}
}
\subfigure[vehicle]{
\label{fig:example1}
\includegraphics[width=.385\columnwidth]{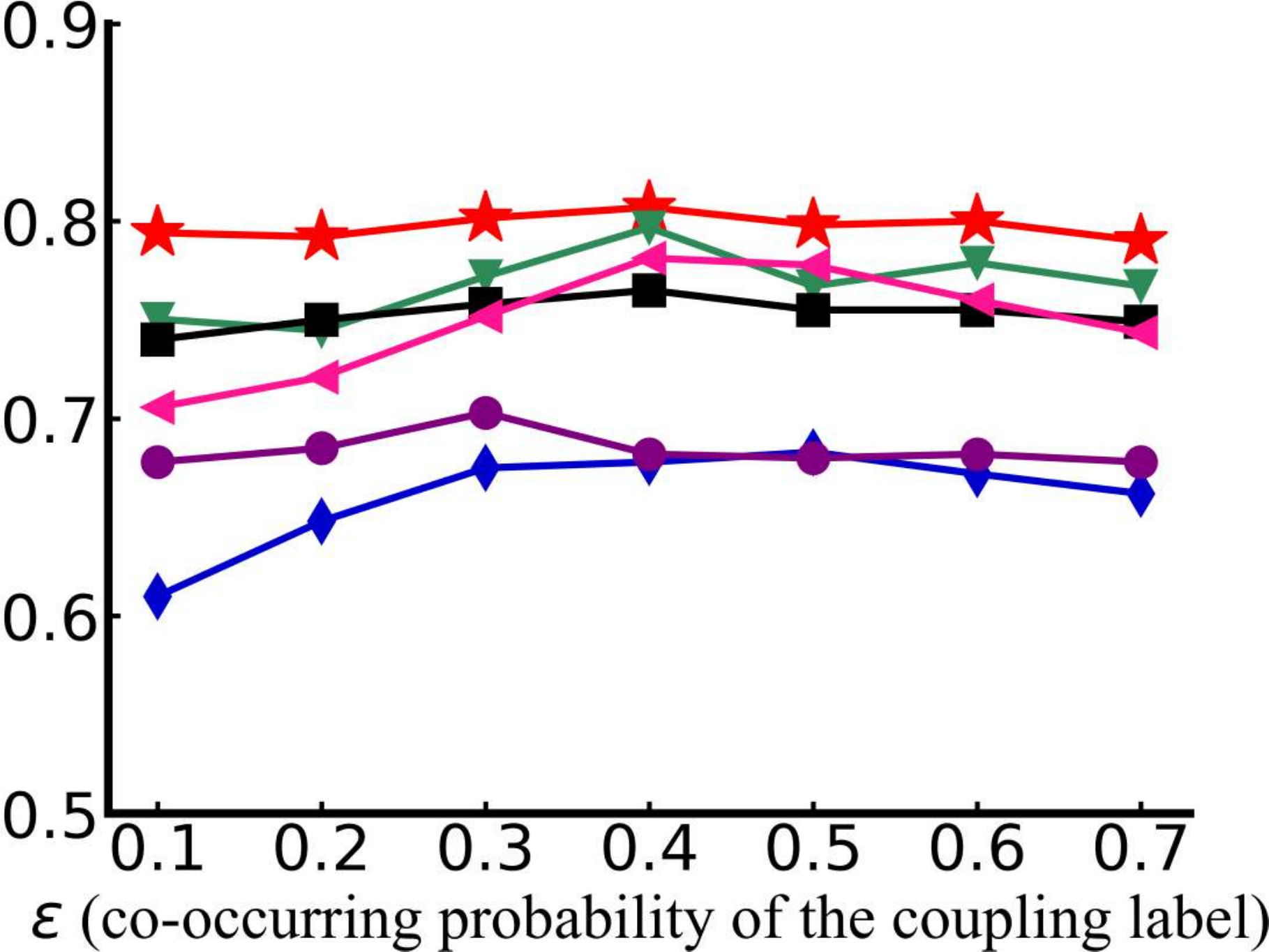}
}
\subfigure[segment]{
\label{fig:example1}
\includegraphics[width=.385\columnwidth]{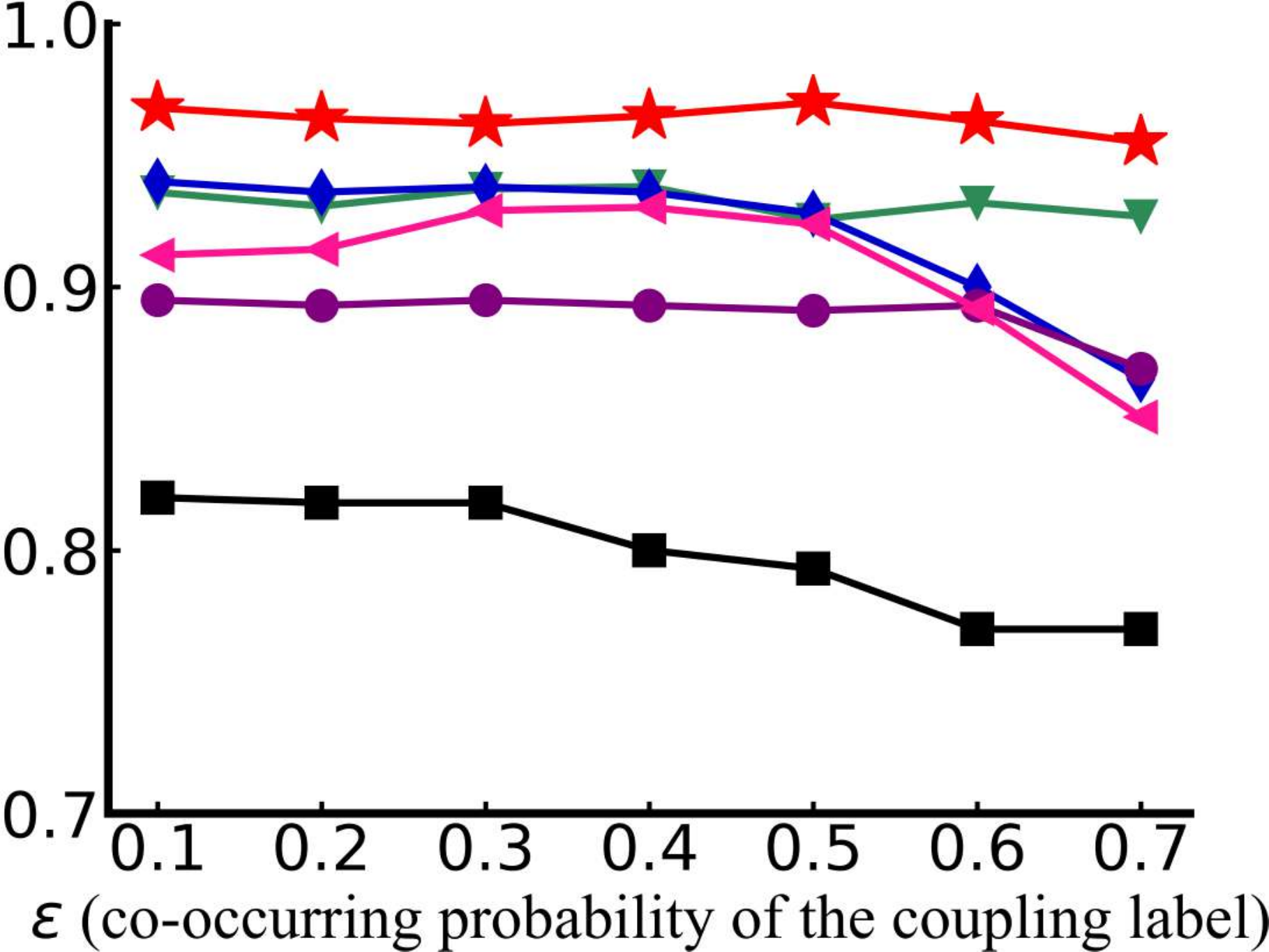}
}
\subfigure[satimage]{
\label{fig:example1}
\includegraphics[width=.385\columnwidth]{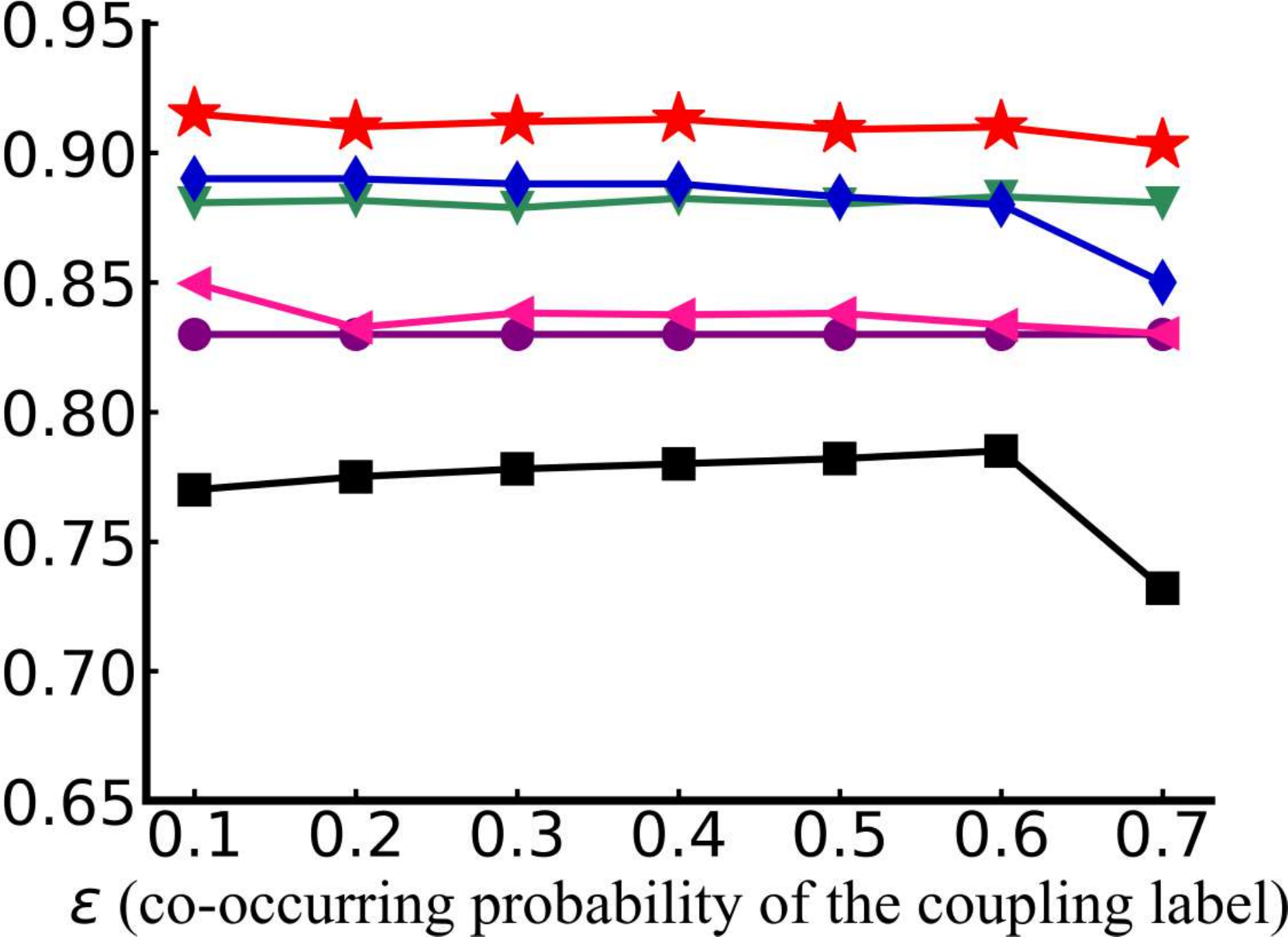}
}
\caption{Test accuracy of each comparison method changes as $\epsilon$ (co-occurring probability of the coupling label) increases from 0.1 to 0.7 (with 100\% partially labeled examples $[p = 1]$ and one false positive candidate label $[r = 1]$).}
\label{coexist}
\end{figure*}

\section{Experiment}
\subsection{Datasets}
We conducted experiments
on 
both controlled synthetic PL datasets
and a number of real-world PL datasets.

The synthetic datasets are generated from four UCI datasets, 
{\em ecoli, vehicle, segment} and {\em satimage},
which have 8, 4, 7 and 7 classes, 
and 336, 846, 2310 and 6,345 examples, respectively. 
From each UCI dataset, we generated synthetic PL datasets using three controlling parameters $p, r$ and $\epsilon$,
following the controlling protocol in previous works \cite{wu2018towards,Xu2019Enhancement,Feng2019Retraining}.
Among the three parameters, $p$ controls the proportion of instances that have noise candidate labels,
$r$ controls the number of false positive labels,
and $\epsilon$ controls the probability of a specific false positive label co-occurring with the true label.
Under different parameter configurations,
multiple PL variants can be generated from each UCI dataset.
In particular, we considered two settings. 
In the first setting, we consider random noise labels 
with the following three groups of configurations:
(I) $r = 1$, $p \in \{0.1, 0.2, \cdot \cdot \cdot, 0.7\}$;
(II) $r = 2$, $p \in \{0.1, 0.2, \cdot \cdot \cdot, 0.7\}$;
(III) $r = 3$, $p \in \{0.1, 0.2, \cdot \cdot \cdot, 0.7\}$.
In the second setting, we consider the target 
label-dependent noise labels with the following configuration:
(IV) $p = 1$, $r = 1$, $\epsilon \in \{0.1, 0.2, \cdot \cdot \cdot, 0.7\}$.
In total, this provides us 112 (28 configurations $\times$ 4 UCI datasets) synthetic PL datasets.

\begin{table}[t!]
	\centering
	\caption{Characteristics of the real-world PL datasets.}\smallskip
	\setlength{\tabcolsep}{5pt}
    \vspace{-0.3cm}
	\label{real_data}
	\smallskip\begin{tabular}{l|c|c|c|c}
				\hline
				{Dataset}      &{\#Example}       &{\#Feature}      &{\#Class}
                                      &{avg.\#CLs}      \\ \hline
                {FG-NET}        &1,002   &262    &78     &7.48       \\ \hline
                {Lost}          &1,122   &108    &16     &2.23  \\ \hline
                {MSRCv2}        &1,758   &48     &23     &3.16  \\ \hline
                {BirdSong}      &4,998   &38     &13     &2.18 \\ \hline
                {Yahoo! News}   &22,991  &163    &219    &1.91  \\ \hline
			\end{tabular}
\end{table}

We used five real-world PL datasets that
are collected from several application domains,
including FG-NET \cite{panis2014overview} for facial age estimation, Lost \cite{cour2011learning},
Yahoo! News \cite{guillaumin2010multiple} for automatic face naming in images or videos,
MSRCv2 \cite{dietterich1994solving} for object classification, and BirdSong \cite{briggs2012rank} for bird song
classification.
The characteristics of the real-world PL datasets 
are summarized in Table \ref{real_data}.

\subsection{Comparison Methods}
We compared the proposed MGPLL approach with the following PL methods,
each configured with the suggested parameters according to the respective literature:
\begin{itemize}[leftmargin=*]
\item PL-KNN \cite{hullermeier2006learning}:
A k-NN based method which makes prediction with weighted voting.
\item PL-SVM \cite{nguyen2008classification}:
A maximum-margin based method which maximizes 
		the classification
margin between candidate and non-candidate class labels.
\item CLPL \cite{cour2011learning}:
A convex optimization based method for partial label learning.
\item PALOC \cite{wu2018towards}:
An ensemble method which trains multiple binary classifies with the one-vs-one decomposition
strategy and makes prediction by consulting all binary classifies.
\item SURE \cite{Feng2019Retraining}:
A self-training based method which learns a confidence matrix of candidate labels
with a maximum infinity norm regularization and trains the prediction model over the learned label confidence matrix.
\end{itemize}

\subsection{Implementation Details}
The proposed MGPLL model has five component networks,
all of which are designed as multilayer perceptrons with Leaky ReLu activation for the middle layers.
The noise label generator is a four-layer network with sigmoid activation in the output layer.
The conditional data generator is a five-layer network with tanh activation in the output layer,
while batch normalization is deployed in its middle three layers.
The predictor is a three-layer network with softmax activation in the output layer.
Both the noise label discriminator and the data discriminator are three-layer networks 
without activation in the output layer.
The RMSProp \cite{tieleman2012lecture} optimizer is used in our implementation and 
the mini-batch size \emph{m} is set to 32.
We selected the hyperparameters $\alpha$, $\beta$ and $\gamma$ from \{0.001, 0.01, 0.1, 1, 10\}
based on the classification loss value $\mathcal{L}_c$ in the training objective function;
that is, we chose 
their
values that lead to the smallest training $\mathcal{L}_c$ loss.


\begin{table}[t!]
	\centering
	\caption{Win/tie/loss counts of pairwise t-test (at 0.05 significance level) between MGPLL and each comparison approach.}\smallskip
	\vspace{-0.3cm}
	\setlength{\tabcolsep}{3.5pt}
    \label{tablest}
	\smallskip\begin{tabular}{l|cccc}
			\hline
                             & \multicolumn{4}{c}{MGPLL vs\ --} \\
                                            &SURE &PALOC &CLPL &PL-SVM  \\ \hline
                varying $p$ $[r = 1]$            &18/7/3      &22/6/0        &24/4/0       &24/4/0         \\
                varying $p$ $[r = 2]$            &16/9/3      &19/9/0        &21/7/0       &22/6/0         \\
                varying $p$ $[r = 3]$            &14/12/2     &18/10/0       &20/8/0       &23/5/0         \\
                
                varying $\epsilon$ $[p, r = 1]$  &15/13/0     &18/10/0       &18/10/0      &21/7/0         \\
            \hline
                              Total               &\textbf{63/41/8}  &\textbf{77/35/0}    &\textbf{83/29/0}    &\textbf{90/22/0}   \\
            \hline
		\end{tabular}%
\end{table}

\subsection{Results on Synthetic PL Datasets}
We conducted experiment on two types of synthetic PL datasets generated from the UCI datasets,
with random noise labels
and target label-dependent noise labels, respectively.
For each PL dataset, ten-fold cross-validation is performed and the average test accuracy results are recorded.
First we study the comparison results over the PL datasets with target label-dependent noise labels
under the PL configuration setting IV.
In this setting, a specific label is selected as the coupled label that co-occurs with the ground-truth label
with probability $\epsilon$,
and any other label can be randomly chosen as a noisy label with probability $1 - \epsilon$.
Figure \ref{coexist} presents the comparison results for the configuration setting IV,
where $\epsilon$ increases from 0.1 to 0.7 with $p = 1$ and $r = 1$.
From Figure \ref{coexist} we can see that the proposed MGPLL produces promising results.
It consistently outperforms the other methods across different $\epsilon$ values on three datasets,
while achieving
remarkable gains on {\em segment} and {\em satimage}.
We also 
conducted experiments on the PL datasets with random noise labels produced
under PL configuration settings I, II and III, 
while MGPLL (with noise label generator $G_n(\epsilon)$) achieves similar positive comparison results as above. 
Due to the limitation of space, 
instead of including the comparison figures, 
we summarize the comparison results below with statistical significance tests.

To statistically 
study the significance of the performance gains achieved by MGPLL over the other comparison methods,
we conducted pairwise t-test at 0.05 significance level based on the comparison results of ten-fold cross-validation
over all the 112 synthetic PL datasets obtained for all different configuration settings.
The detailed win/tie/loss counts between MGPLL and each comparison method are reported in Table \ref{tablest},
from which we have the following observations:
(1) MGPLL achieves superior or at least comparable performance over PALOC, CLPL, and PL-SVM 
in all cases,
which is not easy given the comparison methods have different strengths across different datasets.
(2) MGPLL significantly outperforms 
PALOC, CLPL, and PL-SVM in 
68.7\%, 74.1\%, and 80.3\% 
of the cases respectively, and produces ties in the remaining cases.
(3) MGPLL significantly outperforms SURE in 56.2\% of the cases while achieves comparable performance in 36.6\%,
and is outperformed by SURE in only remaining 7.1\% of the cases.
(4) 
On the PL datasets with 
target label-dependent noise labels, we can see that 
MGPLL significantly outperforms 
SURE, PALOC, CLPL, and PL-SVM in 
53.5\%, 64.2\%, 64.2\%, and 75.0\%,
of the cases respectively.
(5) It is worth noting that MGPLL is never significantly outperformed by any comparison methods
on datasets with label-dependent noise labels.
In summary, these results on the controlled PL datasets clearly demonstrate the effectiveness of MGPLL 
for partial label learning under different settings.

\begin{table*}[th!]
	\centering
	\caption{Test accuracy (mean$\pm$std) of each comparison method on the real-world PL datasets.
$\bullet/$$\circ$ indicates whether MGPLL is statistically superior$/$inferior to the comparison algorithm on each dataset (pairwise t-test at 0.05 significance level).}\smallskip
    \vspace{-0.3cm}
	\label{Comparison_result}
	\smallskip\begin{tabular}{l|l|l|l|l|l|l}
				\hline
				  &MGPLL     &SURE      &PALOC     &CLPL       &PL-SVM      &PL-KNN \\ \hline
                FG-NET       &0.079$\pm$0.024             &0.068$\pm$0.032              &0.064$\pm$0.019
                &0.063$\pm$0.027             &0.063$\pm$0.029              &0.038$\pm$0.025$\bullet$         \\
                FG-NET(MAE3) &0.468$\pm$0.027             &0.458$\pm$0.024               &0.435$\pm$0.018$\bullet$
                &0.458$\pm$0.022             &0.356$\pm$0.022$\bullet$     &0.269$\pm$0.045$\bullet$         \\
                FG-NET(MAE5) &0.626$\pm$0.022             &0.615$\pm$0.019               &0.609$\pm$0.043$\bullet$
                &0.596$\pm$0.017$\bullet$    &0.479$\pm$0.016$\bullet$     &0.438$\pm$0.053$\bullet$         \\
                Lost&0.798$\pm$0.033             &0.780$\pm$0.036$\bullet$     &0.629$\pm$0.056
                &0.742$\pm$0.038$\bullet$    &0.729$\pm$0.042$\bullet$     &0.424$\pm$0.036$\bullet$         \\
                MSRCv2       &0.533$\pm$0.021             &0.481$\pm$0.036$\bullet$     &0.479$\pm$0.042$\bullet$
                &0.413$\pm$0.041$\bullet$    &0.461$\pm$0.046$\bullet$     &0.448$\pm$0.037$\bullet$         \\
                BirdSong     &0.748$\pm$0.020             &0.728$\pm$0.024$\bullet$     &0.711$\pm$0.016$\bullet$
                &0.632$\pm$0.019$\bullet$    &0.660$\pm$0.037$\bullet$     &0.614$\pm$0.021$\bullet$         \\
                Yahoo! News  &0.678$\pm$0.008             &0.644$\pm$0.015$\bullet$     &0.625$\pm$0.005$\bullet$
                &0.462$\pm$0.009$\bullet$    &0.629$\pm$0.010$\bullet$     &0.457$\pm$0.004$\bullet$         \\
                \hline
			\end{tabular}
\end{table*}
\begin{table*}[th!]
	\centering
	\caption{Comparison results of MGPLL and its five ablation variants.}\smallskip
    \vspace{-0.3cm}
	\label{ablation_result}
	\smallskip\begin{tabular}{l|l|l|l|l|l|l}
				\hline
				  &MGPLL            &CLS-w\//o-advn           &CLS-w\//o-advx
                    &CLS-w\//o-g    & CLS-w\//o-aux      &CLS \\ \hline
                FG-NET       &0.079$\pm$0.024             &0.061$\pm$0.024              &0.072$\pm$0.020
                &0.068$\pm$0.029             &0.076$\pm$0.022              &0.057$\pm$0.016                 \\
                FG-NET(MAE3) &0.468$\pm$0.027             &0.430$\pm$0.029              &0.451$\pm$0.032
                &0.436$\pm$0.038             &0.456$\pm$0.033              &0.420$\pm$0.420                  \\
                FG-NET(MAE5) &0.626$\pm$0.022             &0.583$\pm$0.055              &0.605$\pm$0.031
                &0.590$\pm$0.045             &0.612$\pm$0.044              &0.570$\pm$0.034                  \\
                Lost      &0.798$\pm$0.033             &0.623$\pm$0.037              &0.754$\pm$0.032
                &0.687$\pm$0.026             &0.782$\pm$0.043              &0.609$\pm$0.040                  \\
                MSRCv2       &0.533$\pm$0.021             &0.472$\pm$0.030              &0.480$\pm$0.038
                &0.497$\pm$0.031             &0.526$\pm$0.036              &0.450$\pm$0.037                  \\
                BirdSong     &0.748$\pm$0.020             &0.728$\pm$0.010              &0.732$\pm$0.011
                &0.716$\pm$0.011             &0.742$\pm$0.024              &0.674$\pm$0.016                 \\                
                Yahoo! News  &0.678$\pm$0.008             &0.645$\pm$0.008              &0.675$\pm$0.009
                &0.648$\pm$0.014             &0.671$\pm$0.012              &0.610$\pm$0.015                  \\
                \hline
			\end{tabular}
\end{table*}

\subsection{Results on Real-World PL Datasets}
We compared the proposed MGPLL method with the comparison methods on five real-world PL datasets.
For each dataset, ten-fold cross-validation is conducted,
while the mean test accuracy as well as the standard deviation results
are reported in Table \ref{Comparison_result}.
Moreover, statistical pairwise t-test at 0.05 significance level is 
conducted 
to compare MGPLL with each comparison method
based on the results of ten-fold cross-validation.
The significance results are 
indicated in Table \ref{Comparison_result} as well.
Note that the average number of candidate labels (avg.\#CLs) of FG-NET dataset is quite large,
which causes poor performance for all the comparison methods. 
For better evaluation of this facial age estimation task,
we employ the conventional mean absolute error (MAE) \cite{zhang2016partial} to conduct two extra experiments.
Two extra test accuracies are reported on the FG-NET dataset where a test sample
is considered to be correctly predicted if the difference between the predicted age
and the ground-truth age is less than 3 years (MAE3) or 5 years (MAE5).
From Table \ref{Comparison_result} 
we have the following observations:
(1) Comparing with all the five PL methods, MGPLL consistently produces the best results
on all the datasets, 
with remarkable performance gains in many cases.
For example, MGPLL outperforms the best alternative comparison methods by
5.2\%, 3.4\% and 2.0\% on MSRCv2, Yahoo! News and Birdsong respectively.
(2) Out of the total 35 comparison cases (5 comparison methods $\times$ 7 datasets),
MGPLL significantly outperforms all the comparison methods 
across 77.1\% of the cases, and achieves competitive performance in the remaining 22.9\% of cases.
(3) It is worth noting that the performance of MGPLL is never significantly inferior to any other comparison methods.
These results on the real-world PL datasets again validate the efficacy of the proposed method.

\subsection{Ablation Study}
The objective function of MGPLL contains five loss terms: 
classification loss, 
adversarial loss at the label level, 
adversarial loss at the feature level, 
generation loss
and auxiliary classification loss.
To assess the contribution of each part,
we conducted an ablation study by comparing MGPLL with the following ablation variants:
(1) CLS-w\//o-advn, which drops the adversarial loss at the label level.
(2) CLS-w\//o-advx, which drops the adversarial loss at the feature level.
(3) CLS-w\//o-g, which drops the generation loss.
(4) CLS-w\//o-aux, which drops the auxiliary classification loss.
(5) CLS, which only uses the classification loss by dropping all the other loss terms. 
The comparison results are reported in Table \ref{ablation_result}.
We can see that comparing to the full model,
all five variants produce inferior results in general and have performance degradations to different degrees.
This demonstrates that 
the different components in MGPLL
all contribute to the proposed model to some extend.
From Table \ref{ablation_result}, we can also see that the variant CLS-w\//o-advn has a relatively 
larger performance degradation
by dropping the adversarial loss at the label level,
while the variant CLS-w\//o-aux has a small performance degradation by dropping the auxiliary classification loss.
This makes sense as by dropping the adversarial loss for learning noise label generator, 
the generator can produce poor predictions and seriously impact the label denoising of the MGPLL model. 
This suggests that our non-random noise label generation through adversarial learning 
is a very effective and important component for MGPLL.  
For  CLS-w\//o-aux, as we have already got the classification loss on real data, it is reasonable to see
that the auxiliary classification loss on generated data can help but is not critical.
Overall, the ablation study results suggest that 
the proposed MGPLL is effective.

\section{Conclusion}
In this paper, we proposed a novel multi-level generative model, MGPLL, for partial label learning. 
MGPLL uses a conditional label level generator to model target label dependent non-random noise label appearances,
which directly performs candidate label denoising, 
while using a conditional feature level generator to generate data samples from denoised label vectors. 
Moreover, 
a prediction network is incorporated to predict 
the denoised true label of each instance
from its input features, 
which forms inverse mappings between labels and features, together with the data feature generator.
The adversarial learning of the overall model
simultaneously identifies true labels of the training instances from 
both the observed data features and the observed candidate labels, 
while inducing accurate prediction networks that map input feature vectors to (denoised) true label vectors. 
We conducted extensive experiments on real-world and synthesized PL datasets. 
The proposed MGPLL model demonstrates the state-of-the-art PL performance.


\bibliographystyle{abbrv}
\bibliography{paperbib}

\end{document}